\def\year{2018}\relax
\documentclass[letterpaper]{article} 
\usepackage{aaai18}  
\usepackage{times}  
\usepackage{helvet}  
\usepackage{courier}  
\usepackage{url}  
\usepackage{graphicx}  
\usepackage[ruled,linesnumbered]{algorithm2e}
\usepackage{mathrsfs}
\begin{document}
%
\title{Cross-domain  Human Parsing via Adversarial  Feature and  Label Adaptation }
\author{Si Liu$^{1,4,5}$, Yao Sun$^{1,*}$, Defa Zhu$^{1}$, Guanghui Ren$^{1}$, Yu Chen$^{2}$, Jiashi Feng$^{3}$, Jizhong Han$^{1}$\\
	$^1$  Institute of Information Engineering, Chinese Academy of Sciences\\
	$^2$ JD.com
	$^3$ Department of ECE, National University of Singapore \\
	$4$ Jiangsu Key Laboratory of Big Data Analysis Technology /B-DAT,
	Nanjing University of Information Science \& Technology\\
	$5$ Collaborative Innovation Center of Atmospheric Environment and Equipment Technology \\ Nanjing University of Information Science and Technology, Nanjing, China
	\\
	\{liusi, sunyao, zhudefa,  renguanghui, hanjizhong\}@iie.ac.cn, chenyu6@jd.com, elefjia@nus.edu.sg\\
  $*$ corresponding author}
\maketitle

\begin{abstract}
	Human parsing has been extensively studied recently \cite{yamaguchi2012parsing,xia2017joint} due to its wide applications in many important scenarios. Mainstream fashion parsing models (i.e., parsers) 
	focus on parsing the high-resolution and clean images. However, directly applying the parsers trained on benchmarks of high-quality samples  to a particular application scenario in the wild, e.g., a canteen, airport or workplace, often gives non-satisfactory performance  due to domain shift. In this paper, we explore a new and challenging   cross-domain human parsing problem: taking the benchmark dataset with extensive pixel-wise labeling as the source domain, how to obtain a satisfactory parser on a new  target domain \emph{without  requiring any additional manual labeling? }
	To this end, we propose a novel and  efficient  cross-domain human parsing model to bridge  the cross-domain differences  in terms of visual appearance and environment conditions and fully exploit commonalities across domains. 
	Our proposed model  explicitly learns a feature compensation network, which is specialized for mitigating  the cross-domain differences. A discriminative feature adversarial network is introduced to supervise the feature compensation to effectively reduces the discrepancy between feature distributions of two domains. Besides, our  proposed model also introduces a structured label adversarial network to guide  the parsing results of the target domain to follow the high-order relationships of the structured labels shared across domains. The proposed framework is end-to-end trainable, practical and scalable in real applications. Extensive experiments are conducted where LIP dataset is the source domain and $4$ different datasets including surveillance videos, movies and runway shows without any annotations, are evaluated as  target domains. The results consistently confirm data efficiency and performance advantages of the proposed method for the challenging  cross-domain human parsing problem. 
\end{abstract}

\section{Introduction}

Recently human parsing \cite{Liu2015Matching}  has been receiving increasing  attention owning to its wide applications,  such as person re-identification  \cite{zhao2014learning}, people search \cite{li2017person}, fashion synthesis \cite{zhuyour}.

\begin{figure}[t]
	\begin{center}
		\includegraphics[width=0.9\linewidth]{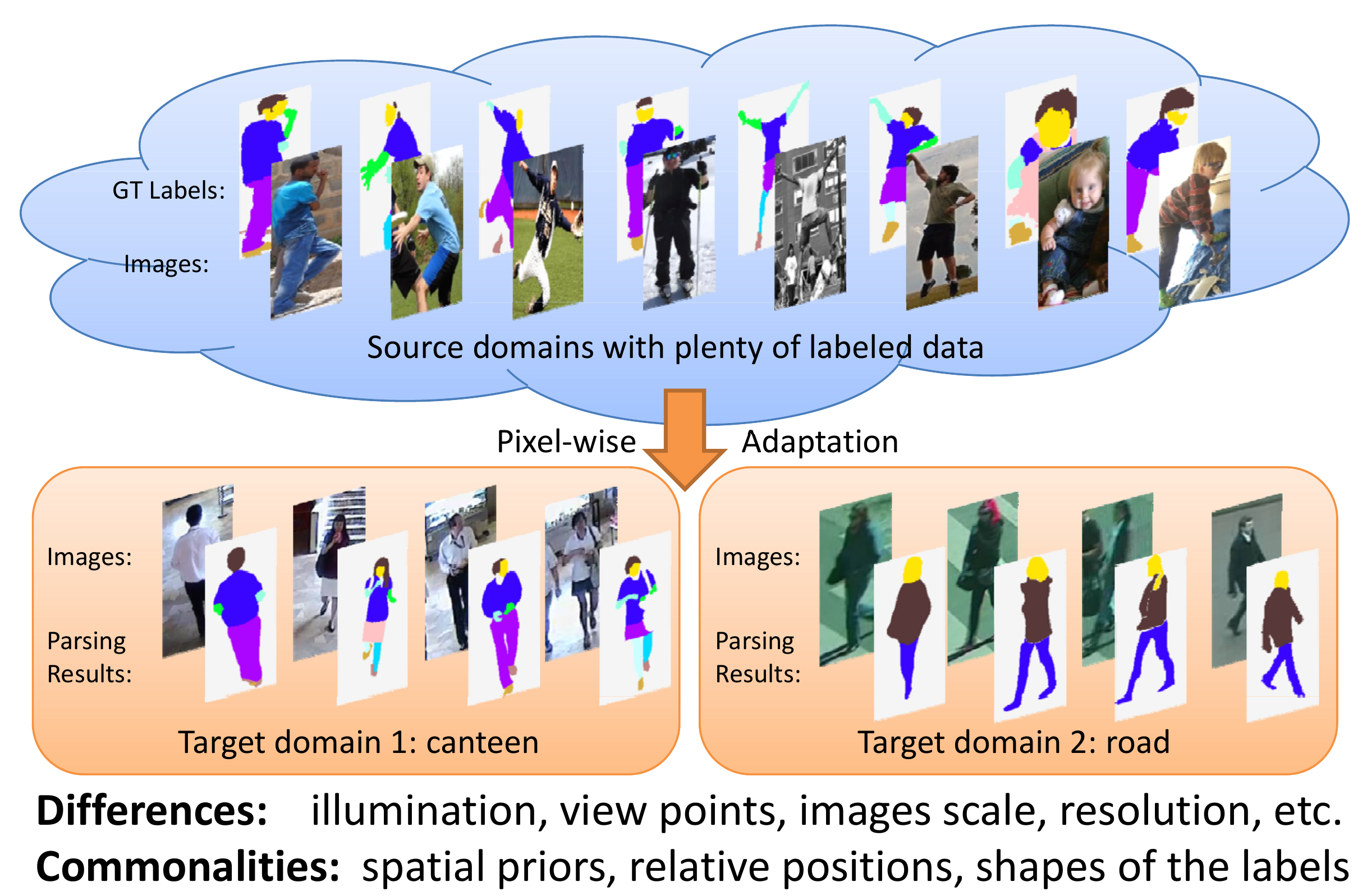}
	\end{center}
	\vspace{-5mm}
	\caption{Cross-domain human parsing: the upper panel is the source domain with a large amount of training data, e.g., the LIP dataset; the lower panel shows the target domain, e.g, canteen and road, without any manual labeling.   }
	\label{fig:first_fig}
	\vspace{-6mm}
\end{figure}

Existing human parsing algorithms can be divided into following two categories. The first one  is \textbf{constrained human parsing}. More specifically,  the clean images of well-posed persons   are collected from some fashion sharing websites, e.g., Chictopia.com, for training and testing.  Representative datasets include  Fashionista \cite{yamaguchi2012parsing} with 685 images, Colorful-Fashion dataset \cite{liu2014fashion} with $2,682$ images and ATR dataset \cite{liang2015deep} with $7,700$ dataset.  Each image in these datasets contains only one person, with relatively simple poses (mostly standing), against relatively clean backgrounds. The human parsers trained in such strictly constrained scenario often fail when applied to  images captured under the real-life, more complicated environments. The second category is \textbf{unconstrained human parsing}. Representative datasets include Pascal human part dataset \cite{chen2014detect} with $3,533$ images and LIP dataset \cite{gong2017look} with $50,462$ images. The images in these dataset present humans with varying clothing appearances, strong articulation, partial (self-) occlusions, truncation at image borders, diverse viewpoints and background clutters. Although they are closer to real environments than the constrained datasets, when applying the human parser trained on these unconstrained datasets to a real application scenario, such as shop, airport, the performance is still worse than the parser trained on that particular scenario even with much less training samples, due to domain shift on visual features.

In this paper, we explore a new \textbf{cross-domain human parsing} problem: taking the unconstrained benchmark dataset with rich pixel-wise labeling as the source domain, how to obtain a satisfactory parser for a totally different target domain without any additional manual labeling?  As shown in Figure \ref{fig:first_fig}, the source domain (shown in the upper panel of Figure \ref{fig:first_fig}) is a set of labeled data. The target domain training set (shown in the lower panel of Figure \ref{fig:first_fig}) is as a set of images without any annotations. We believe investigation on this challenging problem will  push human parsing models   toward more practical applications.

From Figure \ref{fig:first_fig}, we observe the  following  differences and commonality across two domains, e.g., the source domain and the first target domain, canteen. On the one hand, they have very different illumination, view points, image scale, resolution and degree of motion blur etc. For example, the lighting condition in the canteen scenario is much darker than the source domain. On the other hand, the persons to parse from both domains share the intrinsic commonality such as  the high-order relationships among labels (reflecting human body structure) are similar. For example, in both domains, the arms are below the head, but above the legs. Therefore, the cross-domain human parsing problem can be  solved by \emph{minimizing the  differences} of the features and  \emph{maximizing the commonality} of the structured labels.

A typical semantic segmentation network \cite{long2014fully,chen2014semantic}  
is composed of a feature extractor  and a pixel-wise labeler. In this work, we propose to introduce a new and learnable feature compensation network that transforms the features from different domains to a common space where the cross-domain difference can be effectively alleviated. In this way,  the pixel-wise labeler can be readily applied to perform parsing on the compensated features. The feature compensation network is trained under the joint supervision from  a feature adversarial network and a structured label adversarial network. More specifically, the feature  adversarial network serves as a supervisor and provides guidance on the feature compensation learning  like the discriminator of the Generative Adversarial Network (GANs) \cite{goodfellow2014generative,radford2015unsupervised}. It is trained to differentiate target and compensated  source feature  representations. 
Similarly, the structured label adversarial network differentiates the groundtruth structural labels  and the predicted target domain labels. Supervised by these two level information, the cross-domain shift issues can be effectively addressed. We evaluate our approach using  LIP \cite{gong2017look} as  source domain and $4$ datasets 
as target domains. Extensive experiments demonstrate
the effectiveness of our method on all domain shifts adaptation tasks.

The contributions of the paper can be summarized as following. Firstly, we are the first to explore the cross-domain human parsing problem. 
Since  no manual labeling in the target domain is needed,  it is very practical. Secondly, we propose a cross-domain human parsing framework with the novel feature adaptation and structured label adaptation network. It is the first cross-domain work to consider both feature invariance and label structure regularization. 
Thirdly, we will release the source code of our implementation to the academic  to facilitate future studies. 

\vspace{-0.3cm}

\section{Related Work}

Human parsing and cross-domain feature transformation have been studied for decades. However, they generally develop independently. There are few works consider solving the cross-domain human parsing by considering these directions jointly. In this section, we briefly review recent techniques on human parsing as well as feature adaption.  

\textbf{Human parsing:}
Yamaguchi \emph{et al.}     \cite{yamaguchi2013paper}  tackle the clothing parsing  problem using a retrieval based approach. 
Simo-Serra \emph{et al.}  \cite{simo2014high} propose a Conditional Random Field (CRF) model that is able to leverage many different image features.
Luo \emph{et al.} \cite{luo2013pedestrian}  propose a  Deep Decompositional Network  for parsing pedestrian images into semantic regions.  
Liang \emph{et al.} \cite{xiaodaniccv} propose a Contextualized Convolutional Neural Network to tackle the problem and achieve very impressing results. 
Xia \emph{et al.} \cite{xia2015zoom}  propose the ``Auto-Zoom Net'' for human paring. 
%
Wei \emph{et al.} \cite{wei2016hcp,wei2017stc} propose several weakly supervised parsing methods to reduce the human labeling burden. 
Existing human parsing methods work well in the benchmark datasets. However, when applied in the new application scenarios, the performances are unsatisfactory.   The cross-domain human parsing problem becomes a significant problem for making the  technology practical.

\begin{figure*}[t]
	\begin{center}
		\includegraphics[width=0.7\linewidth]{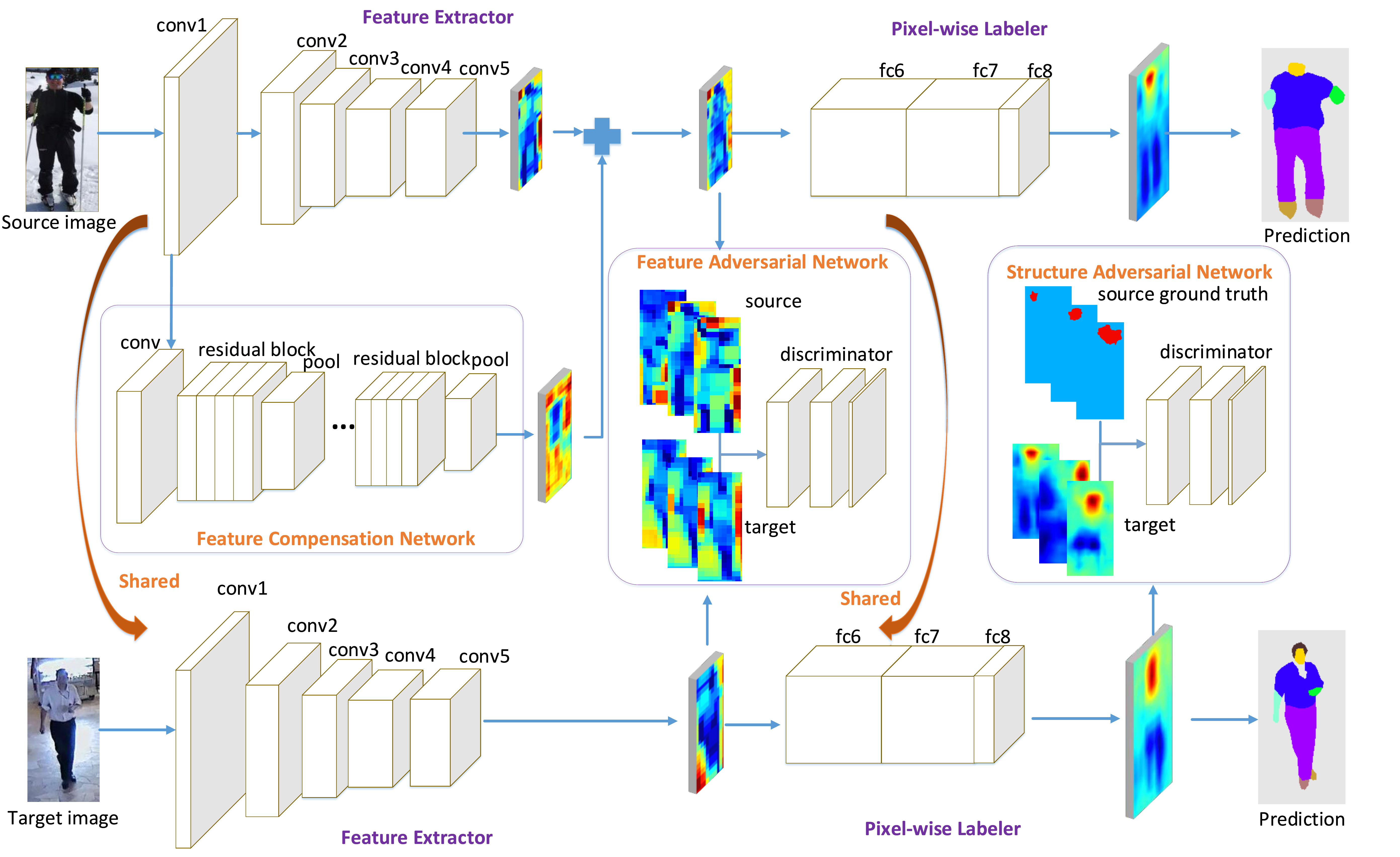}
	\end{center}
	\vspace{-0.5cm}
	\caption{The cross-domain parsing model. It contains 	a  feature adaptation component   to  minimize the feature differences between two domains, and a  structured label adaptation component to maximize the label map commonalities across the domains.  
	}
	\vspace{-6mm}
	\label{fig:framework}
\end{figure*}
\textbf{Feature Adaptation:} 
There have been extensive prior works on domain transfer learning \cite{gretton2009covariate}. Recent works have focused on transferring deep neural network representations from a
labeled source dataset to a target domain where labeled
data is sparse. In the case of unlabeled
target domains (the focus of this paper) the main strategy
has been to guide feature learning by minimizing the
differences between the source and target feature distributions  \cite{ganin2015unsupervised,liu2016coupled,long2015learning}.
Different from  existing feature adaptation methods, we 
explicitly consider the cross-domain differences via a feature compensation network.

\textbf{Structured Label  Adaptation:}  There are few works to consider the label structure adaptation during domain adaption. Some 
pioneer pose estimation works take the geometric constraints of human joints connectivity into consideration. 
For example, Chen \emph{et al.} \cite{chen2017adversarial} propose Adversarial PoseNet, which is a structure-aware convolutional network to implicitly take such priors into account during training of the deep network. 
Chou \emph{et al.}  \cite{chou2017self} 
employ GANs as pose estimator, which enables  learn plausible human body configurations。 
Our proposed cross-domain human parsing method differs from existing domain adaptation methods in that we consider both feature and structured label adaption simultaneously.

%
%
%

\section{Proposed Cross-domain Adaptation Model}

Suppose the source domain images and labels are denoted as $S_x$ and $S_y$ respectively.  The target domain images are represented as $T_x$. 
Typical human parsing models are composed of a feature extractor $E\left(  \cdot  \right)$ and a pixel-wise labeler $L\left(  \cdot  \right)$. However, the parsing model  trained on the source domain does not perform well in the target domain in presence of significant domain shift.

Our proposed cross-domain adaptation model to address this issue is shown in Figure \ref{fig:framework}. It includes a novel feature compensation component supervised by two components, namely adversarial feature adaptation and adversarial structured label adaptation components. 
The feature adaptation component aims to  minimize the feature differences between different domains, while the structured label adaptation is used to maximize the label map  commonalities across the domains. 
Therefore, the whole  model introduces three novel components (shown in purple rectangular) on top of conventional human parsing models: feature compensation network $C\left(  \cdot  \right)$,  feature adversarial network  ${A_f}\left(  \cdot  \right)$ and structured label adversarial network ${A_l}\left(  \cdot  \right)$ to address the cross-domain human parsing problem.
Next, we will introduce the two adversarial learning  components and explain how they help feature compensation to mitigate the domain difference. 

\subsection{Adversarial Feature  Adaptation }

The feature compensation network  and feature adversarial network  collaboratively contribute to the feature  adaptation.  $C\left(  \cdot  \right)$ maps the feature representation of the source domain toward the target domain under the supervision of ${A_f}\left(  \cdot  \right)$. Alternatively updating  them gradually narrows the cross-domain feature differences.  

The \emph{feature compensation network}, as shown in  Figure \ref{fig:framework},  takes as input the extracted features  $E_1(S_x)$ from source domain, where  $E_1\left(  \cdot  \right)$ is a part of the feature extractor 
$E\left(  \cdot  \right)$.
The output $C\left( {{E_1}\left( {{{\cal S}_x}} \right)} \right)$  is the feature differences (shift) between source and target domains. 
$E\left(  \cdot  \right)$ is composed of $5$ convolutional layers of VGG-16 net  \cite{simonyan2014very}, from \emph{conv1} till  \emph{pool5} in VGG-16.The first several layers (from \emph{conv1} till  \emph{pool1})  forms  $E_1\left(  \cdot  \right)$. The structure of the feature compensation network is a ResNet-like \cite{he2016deep} network  with a 7$\times$7 convolution filters and then 6 residual blocks with the identical layout consisting of two 3$\times$3 convolution filters followed by batch-normalization layer and ReLU activation layer. Every three blocks follows a max pool layer and a 3$\times$3 convolution filter to reduce feature maps' sizes.
The result of the feature compensation network is pixel-wisely added to that of the  feature extractor to produce the compensated source domain feature 
$E\left( {{{\cal S}_x}} \right) +C\left( {{E_1}\left( {{{\cal S}_x}} \right)} \right)$.

The \emph{feature adversarial network}  ${A_f}\left(  \cdot  \right)$ is introduced  to  guide the cross-domain feature adaptation. 
Different from traditional adversarial learning models (e.g., vanilla GAN \cite{goodfellow2014generative}) that performs judgment over raw images, our proposed feature adversarial network is defined upon the high-level feature space (\emph{pool5}) which incorporates essential feature information for human parsing. It  can  accelerate the training and inference. The  architecture of ${A_f}\left(  \cdot  \right)$ is composed of the same  fc6-fc7 layers of the  Atrous Spatial Pyramid Pooling (ASPP) scheme in DeepLab \cite{CP2016Deeplab}. Then a convolution layer with 3$\times$3 convolution filters is appended to create a 1-channel probability map,  which is used to calculate the pixel-wise least square feature adversarial loss, like the   local LSGANs \cite{shrivastava2016learning}.

The \emph{optimization} of ${A_f}\left(  \cdot  \right)$ and $C\left(  \cdot  \right)$ are iterative.   More specifically, the objective for updating ${A_f}\left(  \cdot  \right)$  is:
\begin{equation}
\begin{array}{*{20}{l}}
{\mathop {\min }\limits_{{A_f}} \mathcal{E}_{A_f} = \frac{1}{2}{E_{{T_x}\sim{p_{target}}({T_x})}}\left[ {{{\left( {{A_f}\left( {E\left( {{T_x}} \right)} \right) - \textbf{1}} \right)}^2}} \right]}\\
{ + \frac{1}{2}{E_{{S_x}\sim{p_{source}}({{\rm{S}}_x})}}\left[ {{{\left( {{A_f}\left( {\underbrace {{\rm{E}}\left( {{S_x}} \right){\rm{ + }}C\left( {{E_1}\left( {{S_x}} \right)} \right)}_{compensated\;feature}} \right)} \right)}^2}} \right],}
\end{array}
\label{eq:feature_adv}
\end{equation}
where $\bf{1}$ is an all-one tensor.  The feature adversarial network 
adopts the least squares loss function, regressing the feature of the target domain ${E\left( {{T_x}} \right)}$ to $\bf{1}$  while regressing the 
features of the compensated source domain $E\left( {{{\cal S}_x}} \right) +C\left( {{E_1}\left( {{{\cal S}_x}} \right)} \right)$ to $\bf{0}$. 
It  distinguishes the target feature and the compensated source domain feature, while the feature  compensation network tries to transform them into indistinguishable ones.

The learning target of the feature compensation network is to mitigate the difference between source and target features. It is trained by optimizing the following objective function: 
\begin{equation}
\begin{array}{l}
\mathop {\min }\limits_C {{\cal E}_c} = \\
\frac{1}{2}{E_{{S_x}\sim{p_{source}}({{\rm{S}}_x})}}\left[ {{{\left( {{A_f}\left( {\underbrace {{\rm{E}}\left( {{S_x}} \right){\rm{ + }}C\left( {{E_1}\left( {{S_x}} \right)} \right)}_{compensated\;feature}} \right) - {\bf{1}}} \right)}^2}} \right].
\end{array}
\label{eq:compen_fea}
\end{equation}
The target of ${C}\left(  \cdot  \right)$ is to  transform the source domain features to the one similar to target domain by trying to confuse ${A_f}\left(  \cdot  \right)$. 
Or more concretely, the ${C}\left(  \cdot  \right)$ tries to generate features that persuade the ${A_f}\left(  \cdot  \right)$ to predict the feature is from target domain (output binary prediction of \textbf{1}).
It implicitly  maps the source domain features toward the target domain by 
encoding lighting conditions, environment factors.
By iteratively boosting the abilities of ${A_f}\left(  \cdot  \right)$ and $C\left(  \cdot  \right)$  through alternative training, the gap between the two domains are gradually narrowed down.

\vspace{-0.25cm}

\begin{algorithm} [h]
	\SetAlgoNoLine
	\label{alg}
	\caption{Training details of the integrated   cross-domain human parsing framework. }
	\KwIn{Source domain images $S_x$; source domain labels $S_y$;  target domain images $T_x$;  feature extractor $E$;
		feature compensation network  $C$; 	  feature adversarial network  $A_f$;  		 structured label adversarial network $A_l$;  		  pixel-wise labeler $L$;  	 number of training iterations $N$;  	  a constant $K_C$.
	}
	\For{$t = 1, \cdots, N$}
	{
		sample $ \{ S_x^i\}$, $\{ S_y^i\}$, $\{ T_x^i\}$, $i=1, \cdots, n$. \\
		update  $E$, $L$ by minimizing ${\mathcal{P}_{E,L}^{(1)}}$. \\
		update  $C$ by minimizing Equation (\ref{eq:compen_fea}). \\
		update $A_f$ by minimizing Equation (\ref{eq:feature_adv}). \\	
		\If{\mbox{\rm mod}$(t,K_C)==0$}
		{
			update $E$, $L$ by minimizing Equation (\ref{eq:structure_el}). \\	
			update $A_l$ by minimizing Equation (\ref{eq:label_adv}). \\
		}
		update $E$, $L$ by minimizing ${\mathcal{P}_{E,L}^{(2)}}$.\\
	}
\end{algorithm}
\vspace{-2mm}

\vspace{-0.5cm}


%

\subsection{Adversarial Structured Label  Adaptation}

Only feature compensation cannot fully utilize the valuable information about human body structure and leads to suboptimal parsing performance. Therefore, we also propose a structured label adversarial network that learns to capture the commonalities of parsing labels from different domains. Such information is learnable from the source domain data because of the following reasons. Firstly, the labels have very strong spatial priors.  For example, in daily-life scenarios,  the head always lies on the top, while the shoes  appear in the bottom in most cases. 
Moreover, relative positions between the labels are consistent across domains. 
For example, the arms lie on both sides of the body, while the head is at the top of the body.  Finally, the part  shapes of certain labels are basically similar on both domains. For example, the faces are usually round or oval while the legs are often long striped. The pixel-wise labeler and the structured label adversarial network  collaboratively adapt the structured label  prediction.

The \emph{pixel-wise labeler}  is composed of the $fc6$, $fc7$ and $fc8$ layers of    DeepLab \cite{CP2016Deeplab}, which is  a fully convolutional variant of the VGG-16 net  \cite{simonyan2014very}  by modifying the  atrous (dilated) convolutions to increase the field-of-view.   
Depending on the properties of the input, two losses are defined upon the network. 
\begin{itemize}
	\item[1] ${\cal P}_{E,L}^{(1)}$: The pixel-wise cross entropy loss defined upon  the source domain images ${\rm{E}}\left( {{S_x}} \right)$ and ${S_y}$. 
	\item[2] ${\cal P}_{E,L}^{(2)}$: The pixel-wise cross entropy loss defined upon  the compensated source domain features ${{\rm{E}}\left( {{S_x}} \right) + C\left( {{E_1}\left( {{S_x}} \right)} \right)}$ and ${S_y}$. 
\end{itemize}

The \emph{structured label adversarial network}  is used to {distill} the high-order relationships of the labels from the source domain groundtruth pixel-wise labels $S_y$ and \emph{transfer} to guide parsing target domain images.   The architecture of  ${A_l}\left(  \cdot  \right)$ is as follows. LeakyReLU activations and batch normalization are used for all layers except the output. All layers contain stride = 2 convolution filter except the last layer, which just contains  one stride = 1 convolution filter to produce the confidence map. All convolution filter used in the network is 5$\times$5 convolution filter.

The \emph{optimization} is conducted   jointly through a minimax    scheme that alternates between optimizing the parsing    network and the adversarial network.     ${A_l}\left(  \cdot  \right)$ takes either the ground truth label or the prediction parsing result,  and output the probability estimate of  the input is the ground truth (with training target \textbf{1}) or the segmentation network prediction (with training target \textbf{0}). The learning target is: 
\begin{equation}
\begin{array}{*{20}{l}}
{\mathop {\min }\limits_{{A_l}} \mathcal{E}_{A_l}= \frac{1}{2}{E_{{S_y}\sim{p_{source}}({S_y})}}{{\left[ {\left( {{A_l}\left( {{S_y}} \right) - \textbf{1}} \right)} \right]}^2}}\\
{ + \frac{1}{2}{E_{{T_x}\sim{p_{target}}({{\rm{T}}_x})}}\left[ {{{\left( {{A_l}\left( {L\left( {{\rm{E}}\left( {{T_x}} \right)} \right)} \right)} \right)}^2}} \right].}
\end{array}
\label{eq:label_adv}
\end{equation}
The  $A_l$ can help refine the feature extractor  and pixel-wise labeler via:
\begin{equation}
\mathop {\min }\limits_{E,L} {{\cal E}_{E,L}} = \frac{1}{2}{E_{{T_x}\sim{p_{target}}({{\rm{T}}_x})}}\left[ {{{\left( {{A_l}\left( {L\left( {{\rm{E}}\left( {{T_x}} \right)} \right) - {\bf{1}}} \right)} \right)}^2}} \right].
\label{eq:structure_el}
\end{equation}
Both $E\left(  \cdot  \right)$ and $L\left(  \cdot  \right)$ collaboratively confuse $A_f$ to produce the output $\textbf{1}$, which means that the parsing results are drawn from the ground truth labels.

\vspace{-0.25cm}

\subsection{Model Learning and Inference}	

Training details of the integrated   cross-domain human parsing framework are summarized in Algorithm \ref{alg}.  Generally speaking, all the model parameters are alternatively updated. Note that before every update of  $A_l$, the network $E$, $L$, $C$ and $A_f$ are updated for $5$ times.  Experiments show that the different updating scheduling  between $A_l$ and the remaining network facilitate the model convergence.

During inference, the parsing label of the test sample is obtained by 
$L\left( {E\left( {{{\rm{S}}_x}} \right)} \right)$. Note that the feature compensation network, tow adversarial networks are not involved in the inference stage. Therefore, the  complexity of our algorithm is the same with conventional human parsing method. 
\vspace{-0.25cm}

\subsection{Discussions}
In terms of the architecture  of the adversarial networks, we originally tried DCGANs \cite{radford2015unsupervised}. However, we found  it difficult to 
optimize (issue of convergence) and performs not so well. Therefore, we borrow the architecture  Least Squares Generative Adversarial Networks (LSGANs) \cite{mao2016least} to build our adversarial learning networks,  which  adopts least squares loss function for the discriminator. It performs more stable during learning. For the feature adversarial network, the adversarial loss is defined pixel-wisely on the 2-dim feature maps. The \emph{ local LSGANs } structure \cite{shrivastava2016learning} can hence the capacity of the network. The situation is similar for structured label adversarial network. 

%
%

\begin{table}[!t]
	\centering
	\small
	\caption{From LIP to Indoor dataset. ($\%$).}
	\begin{tabular}{c|cccccc}
		\hline
		\hspace{-4mm}		Methods      \hspace{-3mm} & {Avg. acc}    &\hspace{-3mm}  {Fg. acc}  &\hspace{-3mm}  {Avg. pre}  &\hspace{-3mm}   {Avg. rec}  &\hspace{-3mm}  {Avg. F1} \\
		\hline
		\hline
		\hspace{-4mm}		Target Only  \hspace{-3mm} & 89.50	\hspace{-3mm} & 74.53	\hspace{-3mm} & 60.07	\hspace{-3mm} & 59.55	\hspace{-3mm} & 59.75                    \\
		\hline
		\hspace{-4mm}		Source Only   \hspace{-3mm} & 86.84	\hspace{-3mm} & 68.87	\hspace{-3mm} & 51.12	\hspace{-3mm} & 50.97	\hspace{-3mm} & 49.70  	                 \\
		\hline
		\hspace{-4mm}       DANN     \hspace{-3mm} & 88.04	\hspace{-3mm} & 71.74	\hspace{-3mm} & 52.23	\hspace{-3mm} & 50.73	\hspace{-3mm} & 50.50                    \\
		\hline
		\hline
		\hspace{-4mm}    	Feat. Adapt           \hspace{-3mm} & {\bf 88.16} \hspace{-3mm} & 72.56	\hspace{-3mm} & {\bf 53.63}	\hspace{-3mm} & 52.23	\hspace{-3mm} & 51.59                    \\
		\hline
		\hspace{-4mm}    	Lab. Adapt           \hspace{-3mm} & 88.14	\hspace{-3mm} & 72.82	\hspace{-3mm} & 53.21	\hspace{-3mm} & 51.54	\hspace{-3mm} & 50.95                    \\
		\hline
		\hspace{-4mm}    	Feat. +  Lab. Adapt        \hspace{-3mm} & 87.98	\hspace{-3mm} & {\bf 73.86}	\hspace{-3mm} & 50.84	\hspace{-3mm} & {\bf 54.49}	\hspace{-3mm} & {\bf  51.73}                    \\
		\hline
	\end{tabular}%
	\label{tab:compare_baseline_indoor}
	\vspace{-4mm}
\end{table}%

\begin{table}[!t]
	\centering
	\small
	\caption{From LIP to Daily Video dataset. ($\%$).}
	\begin{tabular}{c|cccccc}
		\hline
		\hspace{-4mm}		Methods      \hspace{-3mm} & {Avg. acc}    &\hspace{-3mm}  {Fg. acc}  &\hspace{-3mm}  {Avg. pre}  &\hspace{-3mm}   {Avg. rec}  &\hspace{-3mm}  {Avg. F1} \\
		\hline
		\hline
		\hspace{-4mm}		Target Only   \hspace{-3mm} & 85.96	\hspace{-3mm} & 62.64	\hspace{-3mm} & 58.63	\hspace{-3mm} & 61.07	\hspace{-3mm} & 59.73                    \\
		\hline
		\hspace{-4mm}		Source Only   \hspace{-3mm} & 87.11	\hspace{-3mm} & 63.47	\hspace{-3mm} & 62.05	\hspace{-3mm} & 63.93	\hspace{-3mm} & 62.41  	                 \\
		\hline
		\hspace{-4mm}       DANN     \hspace{-3mm} & 87.56	\hspace{-3mm} & 63.20	\hspace{-3mm} & {\bf 64.28}	\hspace{-3mm} & 62.73	\hspace{-3mm} & 62.84                    \\
		\hline
		\hline
		\hspace{-4mm}    	Feat. Adapt           \hspace{-3mm} & 87.64	\hspace{-3mm} & 64.83	\hspace{-3mm} & 63.95	\hspace{-3mm} & 64.25	\hspace{-3mm} & 63.40                    \\
		\hline
		\hspace{-4mm}    	Lab. Adapt           \hspace{-3mm} & 87.52	\hspace{-3mm} & {\bf 66.53}	\hspace{-3mm} & 62.64	\hspace{-3mm} & 65.62	\hspace{-3mm} & 63.62                    \\
		\hline
		\hspace{-4mm}    	Feat. +  Lab. Adapt       \hspace{-3mm} & {\bf 87.88}	\hspace{-3mm} & 65.87	\hspace{-3mm} & { 64.08}	\hspace{-3mm} & {\bf 65.97}	\hspace{-3mm} & {\bf 64.36}                    \\
		\hline
	\end{tabular}%
	\label{tab:compare_baseline_July}
	\vspace{-4mm}
\end{table}%

\vspace{-0.25cm}

\section{Experiments}

We conduct extensive experiments to evaluate performance of our  model for 4 cross-domain human parsing scenarios. 

\vspace{-0.2cm}
\subsection{Experimental Setting}

\textbf{Source Domain} : We use \emph{LIP} dataset \cite{gong2017look} as the source domain  that contains more than $50, 000$ images with careful pixel-wise annotations of $19$ semantic human parts.  These images are collected from real-world scenarios and the persons present  challenging poses and views, heavily occlusions, various appearances and low-resolutions. The original $19$ labels are merged to $12$ labels or $4$ labels by discarding or combining to be consistent with target domains.

\begin{figure*}[!ht] 
	\begin{center}
		\includegraphics[width=0.75\linewidth]{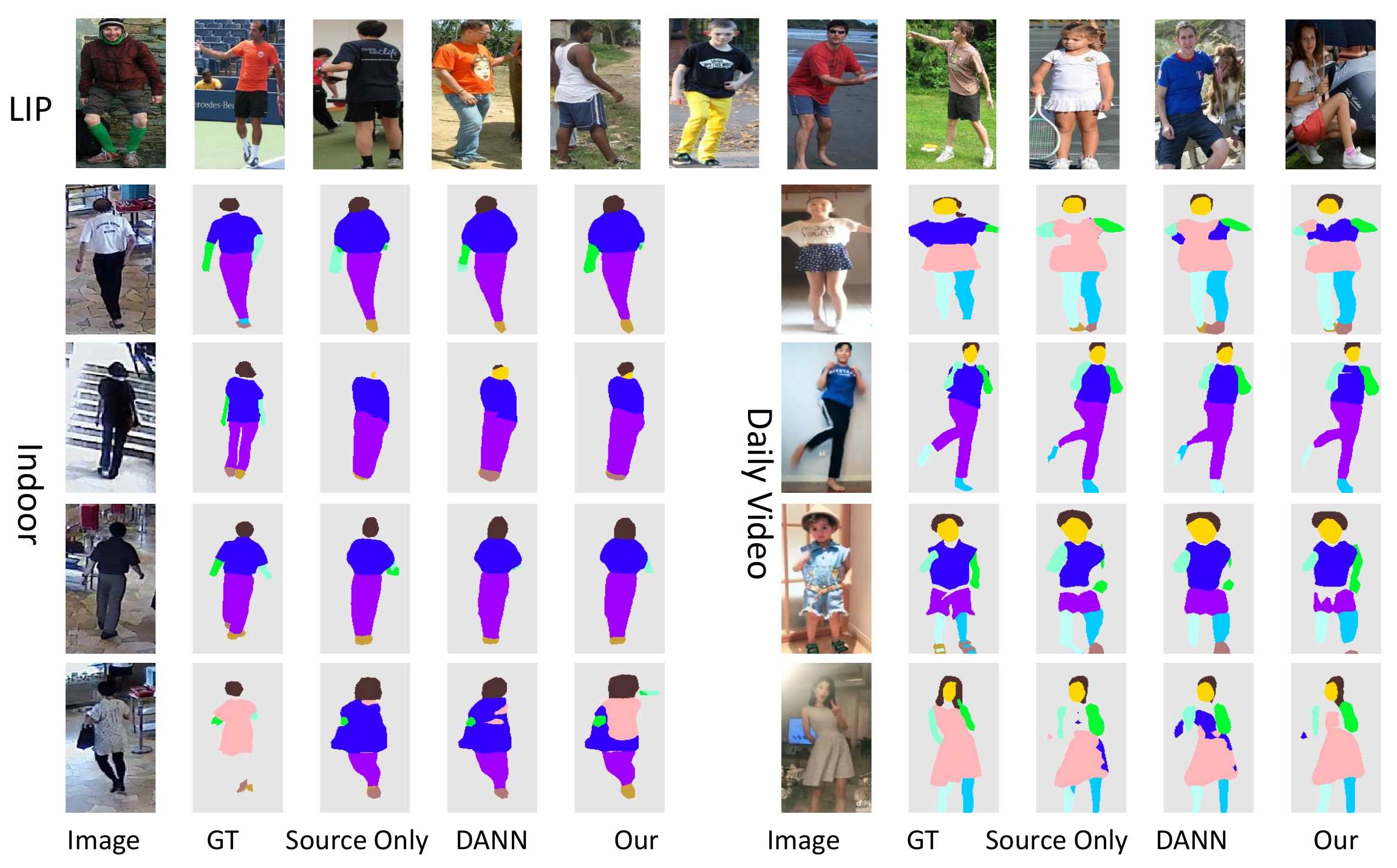}
		\vspace{-2mm}
		\includegraphics[width=0.75\linewidth]{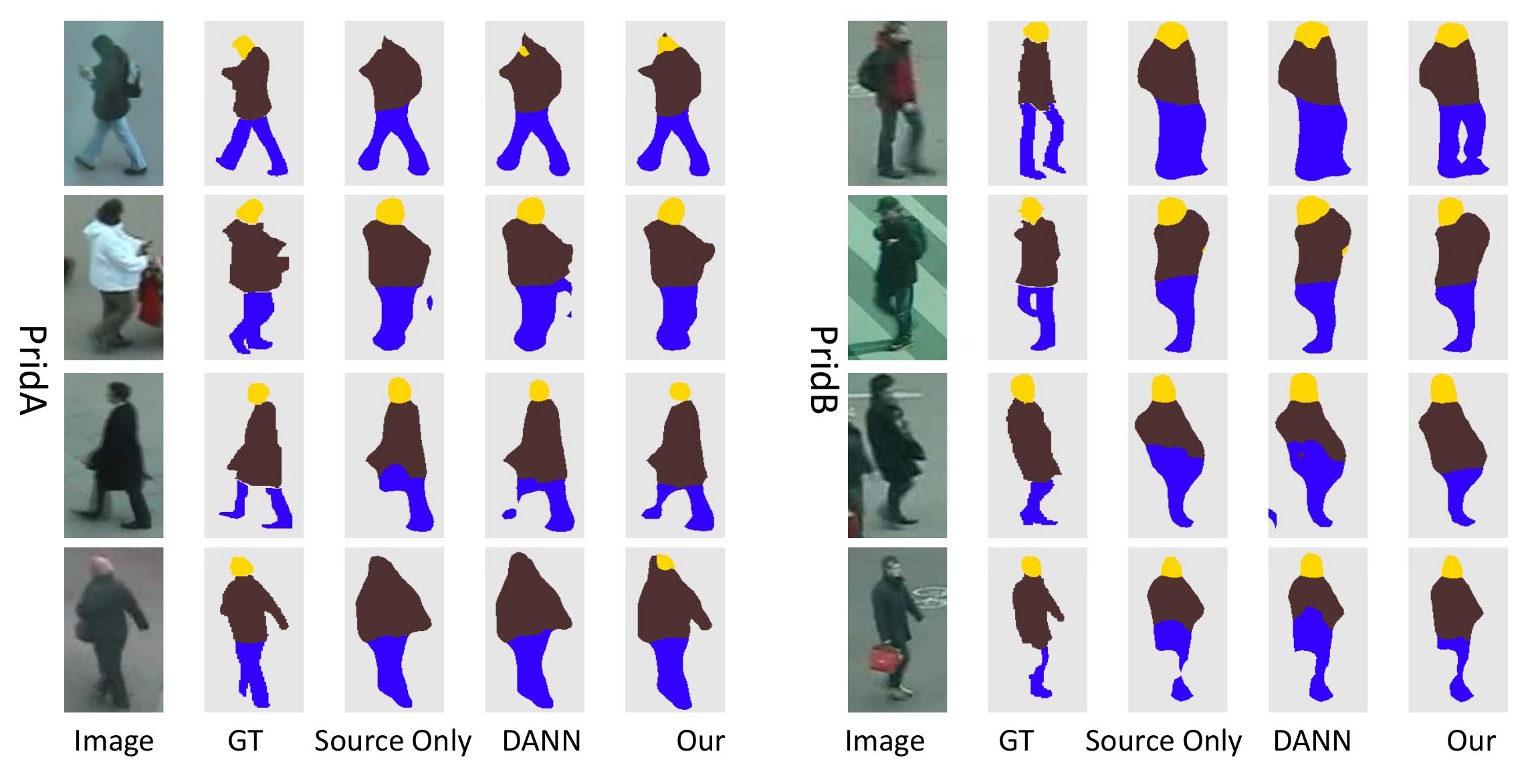}
	\end{center}
	\vspace{-0.5cm}
	\caption{Qualitative Results on 4 target domains.  ``GT'' stands for the groundtruth labels.}
	\label{fig:quantitative_result}
	\vspace{-0.5cm}
\end{figure*}

\textbf{Target Domain}: The following {\em four} target domains are investigated  in this paper. Some example images from these target domains are shown in Figure \ref{fig:quantitative_result}.

\emph{Indoor dataset} \cite{liu2016surveillance} contains $1,900$ labeled images with $12$ semantic human part labels and $15,436$ unlabeled images. The images  are captured in the canteen by surveillance cameras and have dim lights.  

\emph{Daily　 Video dataset} is a newly collected dataset, containing $1,584$ labeled images with $12$ semantic human part labels and $19,964$ unlabeled images. These images are collected from a variety of scenes including shop, road, etc. 


\emph{PridA} and \emph{PridB} datasets are selected from camera view A and camera view B of Person Re-ID 2011 Dataset \cite{roth2014mahalanobis}. Person Re-ID 2011 Dataset consists of images extracted from multiple person trajectories recorded from two different, static surveillance cameras. 


\textbf{Baseline \& Evaluation} 
We compare the proposed method is compared with following baseline methods. 

\noindent {\bf Target Only:} Since all of our target domains have pixel-level  annotations, we train and test the parsing model  directly on the target domains. We take the results, derived from accessing the full supervision, as performance upper bound for the cross-domain parsing models. In the following experiments, the basic model is the same as our feature extraction network and label predicting network. 

\noindent {\bf Source Only:} We apply the model trained on the source domain directly to the target domain, without any fine-tuning on the target domain datasets. It is a valid performance lower bound of the cross-domain methods.

\noindent {\bf DANN:} There are a few works investigating cross-domain learning problems following the adversarial learning strategy. Here, we take the most competitive one proposed in \cite{ganin2016domain}. It resolves the cross-domain problems on classifications. DANN uses an adversary network to make the features extracted from the source domain and target domains  undistinguishable. The feature extraction network are shared for images from both domains. We adapt this method for the human parsing problem.

\begin{table}[!t]
	\centering
	\small
	\caption{From LIP to PridA dataset. ($\%$).}
	\begin{tabular}{c|cccccc}
		\hline
		\hspace{-4mm}		Methods      \hspace{-3mm} & {Avg. acc}    &\hspace{-3mm}  {Fg. acc}  &\hspace{-3mm}  {Avg. pre}  &\hspace{-3mm}   {Avg. rec}  &\hspace{-3mm}  {Avg. F1} \\
		\hline
		\hline
		\hspace{-4mm}		Target Only   \hspace{-3mm} & 89.90	\hspace{-3mm} & 81.44	\hspace{-3mm} & 81.38	\hspace{-3mm} & 82.96	\hspace{-3mm} & 82.12                    \\
		\hline
		\hspace{-4mm}		Source Only   \hspace{-3mm} & 86.10	\hspace{-3mm} & 78.39	\hspace{-3mm} & 72.54	\hspace{-3mm} & 80.60	\hspace{-3mm} & 76.00  	                 \\
		\hline
		\hspace{-4mm}       DANN     \hspace{-3mm} & 86.17	\hspace{-3mm} & {\bf 81.99}	\hspace{-3mm} & 73.51	\hspace{-3mm} & 82.18	\hspace{-3mm} & 76.99                    \\
		\hline
		\hline
		\hspace{-4mm}    	Feat. Adapt           \hspace{-3mm} & 86.63	\hspace{-3mm} & { 81.51}	\hspace{-3mm} & 73.41	\hspace{-3mm} & {\bf 82.88}	\hspace{-3mm} & 77.39                    \\
		\hline
		\hspace{-4mm}    	Lab. Adapt           \hspace{-3mm} & 87.01	\hspace{-3mm} & 79.55	\hspace{-3mm} & 73.74	\hspace{-3mm} & 81.81	\hspace{-3mm} & 77.14                    \\
		\hline
		\hspace{-4mm}    	Feat. +  Lab. Adapt        \hspace{-3mm} & {\bf 87.24}	\hspace{-3mm} & 80.81	\hspace{-3mm} & {\bf 74.76}	\hspace{-3mm} & 82.32	\hspace{-3mm} & {\bf 77.92}                    \\
		\hline
	\end{tabular}%
	\label{tab:compare_baseline_PredA}
	\vspace{-0.5cm}
\end{table}%

\begin{table}[!t]
	\centering
	\small
	\caption{From LIP to PridB dataset. ($\%$).}
	\begin{tabular}{c|cccccc}
		\hline
		\hspace{-4mm}		Methods      \hspace{-3mm} & {Avg. acc}    &\hspace{-3mm}  {Fg. acc}  &\hspace{-3mm}  {Avg. pre}  &\hspace{-3mm}   {Avg. rec}  &\hspace{-3mm}  {Avg. F1} \\
		\hline
		\hline
		\hspace{-4mm}		Target Only   \hspace{-3mm} & 88.50	\hspace{-3mm} & 79.71	\hspace{-3mm} & 79.83	\hspace{-3mm} & 82.28	\hspace{-3mm} & 81.00                    \\
		\hline
		\hspace{-4mm}		Source Only   \hspace{-3mm} & 84.46	\hspace{-3mm} & 80.01	\hspace{-3mm} & 72.85	\hspace{-3mm} & 80.01	\hspace{-3mm} & 75.63  	                 \\
		\hline
		\hspace{-4mm}       DANN     \hspace{-3mm} & 83.91	\hspace{-3mm} & {\bf 83.06}	\hspace{-3mm} & 71.55	\hspace{-3mm} & {\bf 82.83}	\hspace{-3mm} & 75.83                    \\
		\hline
		\hline
		\hspace{-4mm}    	Feat. Adapt           \hspace{-3mm} & 85.63	\hspace{-3mm} & 82.30	\hspace{-3mm} & 74.47	\hspace{-3mm} & { 81.69}	\hspace{-3mm} & 77.28                    \\
		\hline
		\hspace{-4mm}    	Lab. Adapt           \hspace{-3mm} & 84.62	\hspace{-3mm} & 80.54	\hspace{-3mm} & 73.13	\hspace{-3mm} & 80.42	\hspace{-3mm} & 75.89                    \\
		\hline
		\hspace{-4mm}    	Feat. +  Lab. Adapt        \hspace{-3mm} & {\bf 86.26}	\hspace{-3mm} & { 82.39}	\hspace{-3mm} & {\bf 75.20}	\hspace{-3mm} & 81.62	\hspace{-3mm} & {\bf 77.89}                    \\
		\hline
	\end{tabular}%
	\label{tab:compare_baseline_PredB}
	\vspace{-0.5cm}
\end{table}%

\begin{table*}[!ht]
	\caption{F-1 Scores of each category from LIP to Indoor.  ($\%$).}
	\label{tabel:each_category_indoor_state}
	\centering
	\small
	\begin{tabular}{p{2.5cm}|cp{0.6cm}cp{0.6cm}cp{0.6cm}cp{0.6cm}cp{0.6cm}cp{0.6cm}}
		\hline
		\hline
		Methods         & bg	& face	& hair	& U-clothes	& L-arm	& R-arm	& pants	& L-leg	    & R-leg	 	& dress  & L-shoe	& R-shoe		\\
		\hline
		Target Only      & 95.05	& 66.46	& 77.30	& 81.35	    & 50.79	& 50.29	& 80.95  & 38.28	& 39.342 	& 63.15 & 37.285   & 36.68		\\
		\hline
		Source Only      & 94.17	& 58.89	& 59.10	& 77.51	    & {\bf 43.35}	& {\bf 43.39}	& 75.06	 & 35.16	& {\bf 32.53}	    & 26.55	 & 24.11	& 26.54 		\\	
		\hline
		DANN   		& 94.48	& {\bf 61.38}	& 65.26	& 78.41	& 42.01	& 41.74	& 78.83	& 32.84	& 25.53	& 35.56	& 23.76	& 26.19         \\
		\hline
		\hline
		Feat. Adapt	& {\bf 94.53}	& { 58.92}	& 62.99	& 78.27	    & 41.14	& 40.11	& {\bf 79.42}	 & {\bf 41.49}	& 22.90	    & 45.15	 & {\bf 26.53}	& {\bf 27.69 }        \\
		\hline
		Lab. Adapt  			& 94.48	& 57.71	& 63.32	& 78.60	    & 41.20	& 41.22	& 79.06	 & 38.99	& 22.64	    & 45.52	 & 25.90	& 22.70         \\
		\hline
		Feat. +  Lab. Adapt	  		& 94.49	& 56.73	& {\bf 67.86}	& {\bf 78.81}	    & 42.79	& 42.64	& 78.97	 & 36.25	& 22.86	    & {\bf 47.00}	 & 25.32	& 27.00         \\
		\hline
	\end{tabular}
	\vspace{-4mm}
\end{table*}


\begin{table*}[!ht]
	\caption{F-1 Scores of each category from LIP to Daily Video dataset.  ($\%$).}
	\label{tabel:each_category_july_state}
	\centering
	\small
	\begin{tabular}{p{2.5cm}|cp{0.6cm}cp{0.6cm}cp{0.6cm}cp{0.6cm}cp{0.6cm}cp{0.6cm}}
		\hline
		\hline
		Methods         & bg	& face	& hair	& U-clothes	& L-arm	& R-arm	& pants	& L-leg	    & R-leg	 	& dress  & L-shoe	& R-shoe		\\
		\hline
		Target Only      & 94.50	& 69.06	& 57.37	& 68.16	    & 46.33	& 42.37	& 65.01	& 59.97	    & 60.35	    & 67.06	 & 41.85	& 44.72 		\\
		\hline
		Source Only      & 95.15	& 70.28	& 59.54	& 69.91	    & 55.25	& 50.72	& 72.95	& 61.52	    & 61.82	    & 60.32	 & 45.55	& 45.88 		\\	
		\hline
		DANN   		& 95.18	& 70.98	& 58.87	& 71.13	& 54.73	& 50.64	& 73.23	& 61.84	& 61.16	& 64.16	& 46.55	& 45.61         \\
		\hline
		\hline
		Feat. Adapt	  			& 95.35	& {\bf 72.13}	& 55.99	& 73.01	    & 56.55	& 52.38	& 73.08	& 60.60	    & 62.91	    & 61.72	 & {\bf 48.77}	& 48.24         \\
		\hline
		Lab. Adapt  			& 95.37	& 70.99	& {\bf 59.66}	& 72.18	    & 55.94	& 52.33	& 72.76	& 62.68	    & 63.60	    & 63.08	 & 46.51	& 48.31         \\
		\hline
		Feat. +  Lab. Adapt	  		& {\bf 95.38}	& 70.88	& 57.11	& {\bf 73.04}	    & {\bf 57.05}	& {\bf 53.92}	& {\bf 73.34}	& {\bf 64.80}	    & {\bf 64.73}	    & {\bf 64.80}	 & 48.34	& {\bf 48.97}         \\
		\hline
	\end{tabular}
	\vspace{-4mm}
\end{table*}

For ablation studies, we consider three variants of our method,  to evaluate the contribution of each sub-network. {\bf Feat. Adapt:} Our method with the Feature Adversarial network alone. {\bf Lab. Adapt:} Our method with the Structured Label Adversarial network alone. {\bf Feat. + Lab. Adapt:} Our  method with both  Feature Adversarial network and  Structured Adversarial network.

We adopt five popular evaluation metrics, i.e., accuracy, foreground accuracy, average precision, average recall, and average F-1 scores over pixels \cite{yamaguchi2013paper}. All these scores are obtained on the testing sets of the target domains. The annotations of target domains are only used in the ``Target Only'' method.

\textbf{Implementation details}: The feature extractor and the pixel-wise labeler use the DeepLab model, with pre-trained models on PASCAL VOC. The other networks are initialized with ``Normal'' distribution.   

During training of the feature adversarial adaption component, 
``Adam'' optimizer is used with $\beta_ 1 = 0.5$ and $\beta_2 =  0.999$. The learning rate is $1e$-5. When training the structured label adaptation component,
we use ``Adam'' optimizer with $\beta_1 = 0.5$, and $\beta_2 = 0.999$, while the learning rate is $1e$-8. The remaining networks are optimized via  ``SGD'' optimizer with momentum of 0.9, learning rate $1e$-8 and weight decay of 0.0005. The whole framework is trained on PyTorch with a mini-batch size of $10$. The input image size is $241\times 121$. The experiments are done on a single NVIDIA GeForce GTX TITAN X GPU with 12GB memory. The constant $K_C$ is $5$ in our experiment.

\vspace{-0.25cm}


\begin{table}[!t]
	\caption{F-1 Scores of each category from LIP to  PridA dataset.  ($\%$).}
	\label{tabel:each_category_preda_state}
	\centering
	\small
	\begin{tabular}{p{2.5cm}|cp{0.6cm}cp{1.4cm}}
		\hline
		\hline
		Methods         & bg	& head	& U-body	& L-body	   	\\
		\hline
		Target Only      & 93.90	& 76.63	& 81.83	& 76.11 		        \\
		\hline
		Source Only      & 92.06	& 69.05	& 74.49	& 68.38		            \\ 	
		\hline
		DANN        & 92.01	& 71.49	& 75.65	& 68.80	                \\
		\hline
		\hline
		Feat. Adapt	  			& 92.28	& 71.61	& 76.71	& 68.96                 \\
		\hline
		Lab. Adapt  			& 92.78	& 70.24	& 76.11	& 69.43                 \\
		\hline
		Feat. +  Lab. Adapt	  		& {\bf 92.83}	& {\bf 72.01}	& {\bf 76.72}	& {\bf 70.14}                 \\
		\hline
	\end{tabular}
	\vspace{-0.5cm}
\end{table}


\begin{table}[!t]
	\caption{F-1 Scores of each category from LIP to PridB dataset.  ($\%$).}
	\label{tabel:each_category_predb_state}
	\centering
	\small
	\begin{tabular}{p{2.5cm}|cp{0.6cm}cp{1.4cm}}
		\hline
		\hline
		Methods         & bg	& head	& U-body	& L-body		\\
		\hline
		Target Only      & 92.88	& 77.23	& 80.38	& 73.51 		       	\\
		\hline
		Source Only      & 90.75	& 71.79	& 74.56	& 65.41 		        \\	
		\hline
		DANN   		& 90.16	& {\bf 72.73}	& 74.79	& 65.66 		        \\
		\hline
		\hline
		Feat. Adapt	  			& 91.30	& 72.59	& 76.95	& 68.29 		        \\
		\hline
		Lab. Adapt  			& 90.88	& 71.92	& 74.69	& 66.07 		        \\
		\hline
		Feat. +  Lab. Adapt	  		& {\bf 91.59}	& { 72.71}	& {\bf 78.27}	& {\bf 68.99} 		        \\
		\hline
	\end{tabular}
	\vspace{-0.5cm}
\end{table}

\subsection{Quantitative Results}

Table \ref{tab:compare_baseline_indoor} to \ref{tab:compare_baseline_PredB} show the quantitative comparison of the proposed method with baseline methods. The best scores except those performed by ``Target Only'' (upper bound) are shown in black bold.

From these results, we can observe that the ``Feat. +  Lab. Adapt'' method always outperforms about $1\%\sim 2\%$ than the method ``Source Only'' and ``DANN'' in the value ``Avg. F-1'', which verifies the effectiveness of  the proposed cross-domain method. Note that, the ``Avg. F-1'' score of ``Feat. +  Lab. Adapt'' is even higher than those of ``Target Only'' on the Daily Video dataset. We believe the reason is that the number of images in the training set is quite limited in this dataset and our proposed model is effective at transferring useful knowledge to address the sample-insufficiency issue. Besides, ``Feat. Adapt'' performs better than ``Lab. Adapt'' on the dataset Indoor, PridA, and PridB. This is from the fact that the features output by the ``pool5'' layer contain more sufficient characteristics, so the adversary network on these features has more influence on the whole performance.

The detailed ``F-1'' scores of each category are shown in Table \ref{tabel:each_category_indoor_state} $\sim$ \ref{tabel:each_category_predb_state}, which verify the effects of our method.

\vspace{-0.15cm}

\subsection{Qualitative Results}

Some qualitative comparisons on the four target domains are shown  in Figure \ref{fig:quantitative_result}.

For the dataset Indoor, back-view persons appear more frequently, and the illuminations are poor. Therefore, the predictions of left and right arms/shoes are often incorrect, and the hairs may be mis-predicted  as backgrounds as well. For the persons in the 1st and 3rd rows of the dataset Indoor, the left and right arms are confused by ``Source Only''. The DANN  performs slightly better, but our model is able to predict the left and right arms correctly. The hair of the second person is missed in both the ``Source Only'' and ``DANN'' methods, due to the dim lights of the image. The dress of the 4th person looks smaller because the camera is much higher than the person. So ``Source Only'' and ``DANN'' methods wrongly predict them as upper clothes.

For the dataset Daily Video, cameras are put at general positions but the poses of persons are more challenging. People usually appear in frontal view, but they are often moving fast, e.g. the 2nd person, or in nonuniform illuminations, e.g. 3th and 4th persons. In these cases, the proposed model performs better, benefiting from  the structure adversary network. Our method also performs better in predicting the classes of clothes, e.g. the 1st person.

The resolution of the dataset PridA and PridB are very low. As shown in Figure \ref{fig:quantitative_result}, our model and its variants also win in predicting details of the persons.

\vspace{-0.15cm}

\section{Conclusion}
In this paper, we explored a new cross-domain human parsing problem: making use of the benchmark dataset with extensive labeling, how to  build a human parsing for a new scenario without additional labels. 
To this end, an adversarial feature and structured label adaptation method were developed  to learn to minimize the cross-domain feature differences and maximize the label commonalities across the two domains.  In  future,  we plan to explore unsupervised domain adaptation when the target domain are unsupervised  videos. The videos provide rich temporal context and can facilitate cross-domain adaptation. Moreover, we would like to try other types of GANs, such as WGAN \cite{arjovsky2017wasserstein} in our  network.

\vspace{-0.25cm}

\section*{Acknowledgments}

This work was supported by Natural Science Foundation of China (Grant U1536203, Grant 61572493, Grant 61572493),  the Open Project Program of  the Jiangsu Key  Laboratory of  Big Data Analysis Technology, 
Fundamental theory and cutting edge technology Research Program of Institute of Information Engineering, CAS(Grant No. Y7Z0241102) and  Grant No. Y6Z0021102.

\vspace{-0.25cm}

\bibliographystyle{aaai}
\bibliography{egbib_defa}

\end{document}